\documentclass[runningheads]{llncs}

\usepackage[T1]{fontenc}
\usepackage{graphicx}
\usepackage{amsmath,amssymb}
\usepackage{booktabs}
\usepackage{multirow}
\usepackage{tikz}
\usetikzlibrary{shapes, arrows.meta, positioning, fit, backgrounds, calc}
\usepackage{hyperref}
\hypersetup{
    colorlinks=true,
    linkcolor=blue,
    citecolor=blue,
    urlcolor=blue
}

\begin{document}

\title{MRIComp4Flow: Compression of 3D Brain MRI for Training Multi-Modal Generative Models}

\author{Lisa K. Fischer\inst{1}*\and 
Mykhailo Riabets \inst{1}*\and
Daniel Rueckert \inst{4,5}\and
Benedikt Wiestler \inst{2, 3}\and
Anke Meyer-Baese \inst{6, 7}\and
Sandeep Nagar \inst{2, 3, 7}}
\authorrunning{L. K. Fischer \and M. Riabets et al.}
\institute{Technical University of Munich (TUM), Germany \and
Munich Center for Machine Learning (MCML) \and
AI for Image-Guided Diagnosis and Therapy, TUM, Germany \and
Chair for AI in Healthcare and Medicine, TUM and TUM University Hospital,
Munich, Germany \and
Imperial College London \and
Florida State University \and
Institute for Advanced Study, TUM (TUM-IAS), Germany \\
    * equal contribution \\
    \email{ \{lisa.k.fischer, mykhailo.riabets, sandeep.nagar\}@tum.de}
    }

%TODO remember to fuse cropping before publication

\maketitle

\begin{abstract}
Large-scale multi-modal MRI datasets impose substantial storage and I/O costs, limiting the training of 3D generative models on commodity infrastructure.  While lossy compression is known to preserve accuracy for discriminative segmentation networks, its effect on generative models, which must learn the full data distribution rather than a decision boundary, is unexplored. We study whether standard image codecs can effectively compress semantically rich brain tumor MRI while preserving the fidelity required to train and deploy a 3D MRI generative model. Each 3D volume is compressed with JPEG2000 or a near-lossless JPEG-LS pipeline. Next, a Wavelet Flow Matching model, conditioned on BraTS image sequences (T1n, T1c, T2, T2f), is trained on compressed data, and the resulting models are evaluated on the validation set. At a 20:1 compression ratio, synthesis quality is statistically equivalent to a model trained on uncompressed data within a pre-specified margin ($\Delta$PSNR $<1$\,dB, $\Delta$SSIM $<0.02$; paired TOST $p=[[p]]$): mean PSNR is 27.3\,dB vs.\ 27.0\,dB and mean SSIM is 0.95 vs.\ 0.96 across modalities. Our results indicate that JPEG2000 compression is a practical step toward scalable 3D MRI generative modeling without degrading synthesis quality. The code is available at \hyperlink{https://github.com/lisafis/MRIComp4Flow}{https://github.com/lisafis/MRIComp4Flow}.
% and weights and the code can be found at \hyperlink{https://huggingface.co/MRIComp4Flow}{https://huggingface.co/MRIComp4Flow}.

\end{abstract}
% with one-step inference in ${\sim}0.35$\, seconds per volume.
\section{Introduction}

Multi-modal brain MRI is central to tumor segmentation, treatment planning, and longitudinal monitoring \cite{balakrishnan2019voxelmorph,danese2026flowlet,isensee2021nnu,predict_gbm}. Public benchmarks such as BraTS 2024 \cite{brats24,karargyris2023medperf,menze2015multimodal} provide co-registered non-enhanced T1-weighted (T1n), contrast-enhanced T1-weighted (T1c), T2-weighted (T2w), and FLAIR (T2f) volumes. Training 3D generative models on these data requires repeated random access to full volumes during optimization, making storage bandwidth and disk capacity first-order bottlenecks \cite{Khader2023DM_MRi,ryoo2026performance}. For instance, reducing a 1~TB raw NIfTI dataset by a factor of 20 allows the entire corpus to fit on a local NVMe drive, drastically accelerating data loading during training (See Table \ref{tab:dataset_size}).

Recent Diffusion \cite{friedrich2024wdm} and Flow-based \cite{lipman2023flow,tur2026wfm} generative models achieve high-quality cross-modality translation but typically assume lossless NIfTI archives. Kurmukov et al. \cite{kurmukov2024lossy} show that discriminative 3D segmentation networks tolerate JPEG2000 compression up to 20:1 on BraTS MRI with no loss of accuracy, raising the question of whether deep generative models, which must learn the full data distribution rather than a decision boundary, inherit the same robustness. We ask a practical question: \emph{how far can 3D MRI be compressed before generative model training and inference quality degrade?} Answering this would enable larger cohorts, faster data transfer, and cheaper archival while retaining the anatomical and contrast information that generative models exploit.

We propose \textbf{MRIComp4Flow}, a pipeline that compresses BraTS volumes using widely supported codecs, reconstructs 3D tensors on the fly during training, and trains a unified Wavelet Flow Matching (WFM)~\cite{tur2026wfm} model on the compressed corpus. We compare models trained on uncompressed NIfTI data against models trained on JPEG2000 (ratios 2:1--400:1) and near-lossless JPEG-LS (\texttt{NEAR} levels 1--255) across all four modalities. The contributions of this paper can be summarized as:
\begin{enumerate}
    \item \emph{Compression pipeline:} A slice-wise JPEG2000/JPEG-LS pipeline and compressed data loader for 3D MRI, compatible with WFM training.
    \item \emph{Generative robustness:} The first characterization, to our knowledge, of how generative synthesis quality varies with input compression, showing that WFM tolerates JPEG2000 up to 20:1 and thereby extending prior discriminative model findings~\cite{kurmukov2024lossy} to generative models.
    \item \emph{Comprehensive evaluation:} Quantitative and qualitative synthesis results across nine JPEG2000 ratios and seven JPEG-LS levels, downstream tumor segmentation with a state-of-the-art BraTS2025 \cite{brats25} algorithm, and ablations on the number of sampling steps and the contribution of the informed prior.
\end{enumerate}

\section{Related Work}

\paragraph{Medical image compression:} JPEG2000 and JPEG-LS are established standards in radiological workflows \cite{christopoulos2000jpeg2000,schelling2004jpegls,weinberger2000loco}. JPEG2000 \cite{christopoulos2000jpeg2000} offers embedded rate control via compression ratios and is used in DICOM storage profiles; JPEG-LS \cite{weinberger2000loco} provides efficient near-lossless coding through a maximum pixel error (\texttt{NEAR}) parameter. Prior work on MRI compression \cite{kurmukov2024lossy} has focused on diagnostic fidelity rather than downstream machine learning tasks. In this paper, we therefore explicitly measure the impact of compression through generative synthesis quality, which is sensitive to fine anatomical detail and inter-modality contrast potentially affected by compression.

\paragraph{Multi-modal MRI synthesis:} Conditional generative models, including GANs, diffusion models, and flow-matching methods, synthesize missing MRI contrasts from available sequences \cite{friedrich2024wdm,li2023brats,wyatt2022anoddpm}.
Wavelet Flow Matching (WFM) \cite{tur2026wfm} learns a velocity field in Haar wavelet space from an informed prior (the mean of observed modalities) to the target contrast, enabling accurate synthesis with a single 82M-parameter network for all four BraTS modalities. We adopt WFM as our generative backbone because it is state-of-the-art in terms of speed and quality on BraTS~2024, enabling us to isolate the effect of input compression.

\paragraph{Learning from compressed data:} Training on compressed images is common in natural-image domains but remains underexplored for 3D medical volume synthesis \cite{kurmukov2024lossy}. Our work bridges lossy archival formats and 3D generative modeling by evaluating whether codec-induced errors survive wavelet-domain flow matching or are absorbed during normalization and augmentation.

\section{Method}

\begin{figure}[t]
    \centering
    \resizebox{\textwidth}{!}
    {%
    \begin{tikzpicture}[
        >=Stealth,
        font=\small,
        node distance=0.6cm and 0.7cm,
        databox/.style={
            draw=black!80, rounded corners=3pt, line width=0.8pt,
            fill=blue!10, minimum height=1.1cm, minimum width=2.2cm,
            align=center, inner sep=5pt},
        processbox/.style={
            draw=black!70, rounded corners=3pt, line width=0.6pt,
            fill=orange!10, minimum height=1.1cm, minimum width=2.2cm,
            align=center, inner sep=5pt},
        modelbox/.style={
            draw=black!90, rounded corners=4pt, line width=1.2pt,
            fill=green!15, minimum height=1.3cm, minimum width=3.2cm,
            align=center, inner sep=6pt
        },
        storebox/.style={
            draw=black!70, rounded corners=3pt, line width=0.6pt,
            fill=gray!20, minimum height=1.0cm, minimum width=2.0cm,
            align=center, inner sep=1pt},
        stage/.style={
            draw=black!50, dashed, rounded corners=8pt, line width=1pt,
            inner sep=0.4cm, fill=black!5 },
        stagelabel/.style={ font=\normalsize\bfseries, text=black!80},
        arrow/.style={ ->, line width=0.9pt, draw=black!80},
        dashedarrow/.style={ ->, line width=0.9pt, draw=black!60, dashed}
    ]
    
    % ----- Stage I: Offline compression -----
    \node[databox] (nifti) {BraTS 3D NIfTI\\[-1pt]\scriptsize (T1n, T1c, T2w, T2f)};
    \node[processbox, right=of nifti] (slice) {Axial slice\\[-1pt]\scriptsize (uint16)};
    \node[processbox, right=of slice] (codec) {JPEG2000\\[-1pt]\scriptsize or JPEG-LS};
    \node[storebox, right=of codec] (archive) {Compressed\\[-1pt]\scriptsize archive ($r$)};
    
    \draw[arrow] (nifti) -- (slice);
    \draw[arrow] (slice) -- (codec);
    \draw[arrow] (codec) -- (archive);
    
    % Stage I Bounding Box (drawn behind using fit)
    \node[stagelabel, above=0.2cm of codec] (lbl_offline) {Stage I: Offline Compression};
    \begin{scope}[on background layer]
        \node[stage, right=0.22cm of slice, fit=(nifti)(archive)(lbl_offline)] (stage1box) {};
    \end{scope}
    
    % ----- Stage II: Reconstruction & WFM -----
    \node[processbox, below=2.0cm of nifti] (decode) {Decode\\[-1pt]\scriptsize (.jp2 / .jls)};
    \node[processbox, right=of decode] (stack) {Stack \&\\[-1pt]\scriptsize normalize};
    \node[processbox, right=of stack] (dwt) {3D Haar\\[-1pt]\scriptsize DWT};
    \node[modelbox, right=of dwt] (wfm) {\textbf{WFM}\\[-1pt]\scriptsize (unified)};
    \node[processbox, right=of wfm] (idwt) {IDWT\\[-1pt]\scriptsize (reconstruct)};
    \node[databox, right=of idwt] (output) {Synthesized\\[-1pt]\scriptsize 3D MRI};
    
    \draw[arrow] (decode) -- (stack);
    \draw[arrow] (stack) -- (dwt);
    \draw[arrow] (dwt) -- (wfm);
    \draw[arrow] (wfm) -- (idwt);
    \draw[arrow] (idwt) -- (output);
    
    % Inter-stage routing
    \draw[arrow] (archive.south) -- ++(0,-0.6) -| (decode.north);
    
    % Conditioning Inputs
    \node[databox, below=0.8cm of stack] (cond) {Conditioning\\[-1pt]\scriptsize modalities $\mathcal{C}$};
    \node[processbox, right=0.5cm of cond] (prior) {Informed prior\\[-1pt]\scriptsize (mean in wavelet space)};
    
    \draw[arrow] (cond) -- (prior);
    \draw[arrow] (prior.east) -| (wfm.south);
    
    % Stage II Bounding Box
    \node[stagelabel, above=0.1cm of wfm] (lbl_online) {Stage II: WFM Synthesis \& Reconstruction};
    \begin{scope}[on background layer]
        \node[stage, right=-0.2cm of slice, fit=(decode)(output)(cond)(lbl_online)] (stage2box) {};
    \end{scope}
    
    \node[font=\scriptsize, text=black!70, align=center, below=0.2cm of wfm]{1-step flow matching \\ at inference};
    
    \end{tikzpicture}%
    }
    \caption{Overview of our proposed MRIComp4Flow pipeline.}
    \label{fig:pipeline}
    \vspace{-1.5em}
\end{figure}
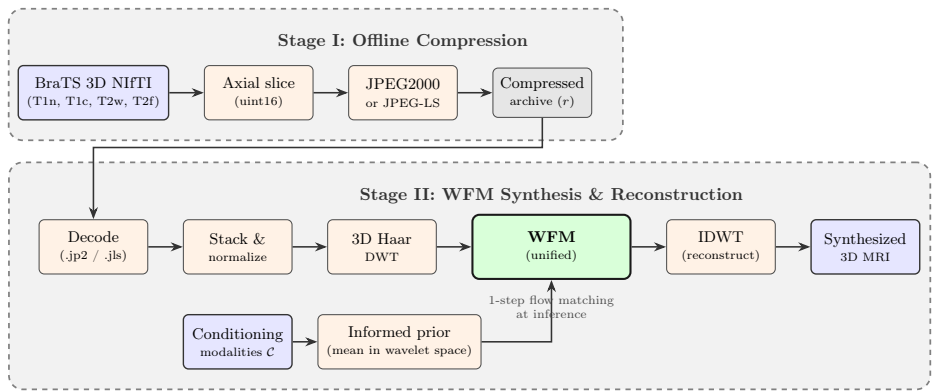

MRIComp4Flow comprises three stages: compression, synthesis, and reconstruction (Fig. \ref{fig:pipeline}). First, raw BraTS NIfTI volumes are compressed offline into per-slice archives. Next, for training a generative model, slices are decoded, stacked into 3D volumes, intensity-normalized, and fed to WFM in Haar wavelet space. Finally, during inference, the model synthesizes missing modalities.

\subsection{3D MRI Compression} 

Given a patient directory with co-registered modalities  $$\{V_m\}_{m \in \mathcal{M}}, \mathcal{M} = \{\mathrm{t1n}, \mathrm{t1c}, \mathrm{t2w}, \mathrm{t2f}\},$$ each volume $V_m \in \mathbb{R}^{H \times W \times D}$ is processed independently. Before compression, voxel intensities are transformed to $[0, \infty)$, cast to uint16, and segmentation masks are excluded from lossy coding. Each axial slice is then encoded either with JPEG2000 (9/7 irreversible transform, \texttt{glymur}) \cite{christopoulos2000jpeg2000} at ratios $r \in [2,400]$, skipping empty background slices, or with near-lossless JPEG-LS (\texttt{NEAR} error; \texttt{NEAR=0} is visually lossless for 16-bit data). The ratio $r$ is the codec's \emph{nominal} per-slice target; the realized whole-dataset reduction is smaller because empty background slices are skipped and container overhead is added. For example, the nominal 20:1 JPEG2000 setting compresses the BraTS training set from $40.37$\,GB to $3.13$\,GB, a $12.9\times$ reduction (Table~\ref{tab:dataset_size}).

% At 20:1, this yields a 20-fold reduction in slice payload compared to raw \texttt{uint16} storage, substantially reducing the large BraTS training dataset size.

The compressed loader decodes all \texttt{.jp2} (or \texttt{.jls}) slices, and inserts each slice at its axial index into a $182 \times 218 \times 182$ buffer. Volumes are clipped to the 0.1st - 99.9th intensity percentiles, min--max normalized to $[0,1]$, spatially padded to $240 \times 240 \times 160$ and center-cropped, then trimmed to $224 \times 224 \times 160$ to match the WFM input geometry.

% This configuration was maintained for consistency with model training, although we want to evaluate whether the intermediate cropping step can be removed.

\subsection{Wavelet Flow Matching (WFM) Backbone}

We used the WFM as a generative model baseline that operates in the 3D Haar wavelet space. Let $\mathrm{DWT}(\cdot)$ denote the 3D discrete wavelet transform producing eight subbands concatenated along the channel dimension. Then, for target modality $y$ conditioned on the modality set $\mathcal{C}$, the informed prior is defined as the mean of the available modalities in wavelet space:
\begin{equation}
    x_{\mathrm{source}} = \frac{1}{|\mathcal{C}|} \sum_{c \in \mathcal{C}} \mathrm{DWT}(x_c).
\end{equation}

\paragraph{Forward process and objective:} Following conditional flow matching, noisy states interpolate between source and target with a $\sigma$-schedule peaking at $t=0.5$:
\begin{equation}
    x_t = (1-t)\,x_{\mathrm{source}} + t\,x_{\mathrm{target}} + \sigma(t)\,\epsilon, \quad \sigma(t) = \sigma_{\max}\sqrt{t(1-t)}.
\end{equation}
A U-Net $f_\theta$ predicts the constant velocity $v = x_{\mathrm{target}} - x_{\mathrm{source}}$ by minimizing
\begin{equation}
    \mathcal{L} = \mathbb{E}_{t,\epsilon}\left[\left\| f_\theta(\tilde{x}_t, c, t, y) - (x_{\mathrm{target}} - x_{\mathrm{source}}) \right\|^2 \right],
\end{equation}
where $c$ stacks conditioning wavelet coefficients and $y$ is a class label over the four modalities. At inference, one Euler step from $t=0$ to $t=1$ followed by $\mathrm{IDWT}(\cdot)$, i.e., inverse $\mathrm{DWT}(\cdot)$ operation, yields the synthesized volume.

% \subsection{Experimental Conditions}

\section{Experiments}

\paragraph{Dataset Details:} We use the BraTS2024 Glioma (GLI) cohort with four co-registered MRI modalities per case. All experiments use BraTS~2024 Glioma cohort ($1{,}621$/$188$ train/val), comprising four co-registered $182 \times 218 \times 182$ modalities (T1n, T1c, T2w, T2f) at $1\,\mathrm{mm}$ isotropic resolution \cite{bratscluster24,brats24,karargyris2023medperf}. Evaluation uses a validation set; for each case and each target modality, the model receives three remaining modalities as conditioning and synthesizes the held-out contrast.

\begin{table}[!ht]
    \centering
    \caption{Storage of BraTS~2024 train set  under both codecs; uncompressed size is 40.37\,GB. Size in GB with a reduction factor relative to uncompressed ($\times$).}
    \label{tab:dataset_size}
    \small
    % \setlength{\tabcolsep}{6pt}
    % \vspace{-0.5em}
    \begin{tabular}{lcc}
    \toprule
    Setting ($r$ / NEAR) & JPEG2000 & JPEG-LS \\
    \midrule
    1   & --                                  & $7.04$ ($5.7\times$)  \\
    2   & $10.66$ ($3.8\times$) & $6.24$ ($6.5\times$)  \\
    5   & $9.44$ ($4.3\times$)  & $5.03$ ($8.0\times$)  \\
    10  & $5.78$ ($7.0\times$) & $4.12$ ($9.8\times$)  \\
    20  & $3.13$ ($12.9\times$) & $3.25$ ($12.4\times$) \\
    50  & $1.31$ ($30.8\times$) & $2.24$ ($18.0\times$) \\
    100 & $0.68$ ($59.4\times$) & --             \\
    200 & $0.35$ ($115.3\times$) & $1.13$ ($35.7\times$) \\
    % 255 & --                    & $0.99$ ($40.8\times$) \\
    400 & $0.21$ ($192.2\times$) & -- \\
    \bottomrule
    \end{tabular}
    \vspace{-1em}
\end{table}

\paragraph{Implementation and Training Details:} For training, all models share an identical configuration: an ${\sim}82M$\, parameter WavUNet (unified 4-class conditioning, 3D Haar wavelets) trained with Adam (lr $10^{-5}$), batch size 4, and $\sigma_{\max}=0.5$. We train 18 unified WFM models with the same hyperparameters, including a Baseline (uncompressed) model using the original BraTS NIfTI training data. Compressed: JPEG2000 at 20:1, reconstructed on-the-fly. Compressed JPEG-LS with near=20, reconstructed on-the-fly. Additional experiments at ratios 1 (JPEG-LS only), 2, 5, 10, 50, 100 (JPEG2000 only), 200, 255 (JPEG-LS only), 400 (JPEG2000 only) follow the same protocol.

\subsection{Results}

Synthesized volumes are saved as NIfTI and compared against the uncompressed ground-truth references. We report PSNR, SSIM (3D, \texttt{skimage}), after per-volume min--max normalization to $[0,1]$. Timing is measured for each subject across all four synthesized modalities. Unless noted otherwise, we report per-subject means across the 188 validation cases with 95\% confidence intervals; baseline-vs-Comp-20 differences are assessed with a paired Wilcoxon signed-rank test and a two-one-sided-tests (TOST) equivalence test at the margins above.

\paragraph{Reconstruction fidelity of compressed inputs:} We first quantify how faithfully each codec reconstructs the \emph{input} slices before any synthesis (Table~\ref{tab:slice_psnr}). JPEG2000 is effectively lossless at low ratios ($\infty$\,dB PSNR, SSIM $=1.0000$ at $r\in\{2,5\}$) and degrades gracefully, still retaining $38.05$\,dB / $0.9741$ SSIM at 20:1. JPEG-LS loses fidelity faster as \texttt{NEAR} grows, dropping from $54.45$\,dB at \texttt{NEAR}$=$1 to $24.70$\,dB at \texttt{NEAR}$=$50; notably, JPEG2000 at the aggressive 100:1 setting ($25.70$\,dB) is comparable to JPEG-LS at \texttt{NEAR}$=$50.

\begin{table}[!htbp]
    \centering
    \caption{Fidelity of the reconstructed compressed slices for JPEG2000 (ratio $r$) and JPEG-LS (\texttt{NEAR}). $\infty$ denotes near-lossless reconstruction.}
    \label{tab:slice_psnr}
    \small
    \vspace{-0.5em}
    \begin{tabular}{lcccc}
        \toprule
        \multirow{2}{*}{Setting} & \multicolumn{2}{c}{JPEG2000} & \multicolumn{2}{c}{JPEG-LS} \\
        \cmidrule(lr){2-3}\cmidrule(lr){4-5}
         & PSNR (dB) & SSIM & PSNR (dB) & SSIM \\
        \midrule
        1   & --       & --       & $54.45$ & $0.9954$ \\
        2   & $\infty$ & $1.0000$ & $50.04$ & $0.9920$ \\
        5   & $\infty$ & $1.0000$ & $43.24$ & $0.9785$ \\
        10  & $46.30$  & $0.9960$ & $37.78$ & $0.9494$ \\
        20  & $38.05$  & $0.9741$ & $32.47$ & $0.9066$ \\
        50  & $30.64$  & $0.8905$ & $24.70$ & $0.7318$ \\
        100 & $25.70$  & $0.7711$ & --      & --       \\
        % 200 & $23.25$  & $0.7711$ & --      & --       \\
        % 400 & $25.70$  & $0.7711$ & --      & --       \\
        \bottomrule
    \end{tabular}
    \vspace{-1em}
\end{table}

\paragraph{Synthesis Quality:} Table~\ref{tab:jpeg2000_compression} reports per-modality synthesis quality across all JPEG2000 ratios. The key result is that input compression does \emph{not} degrade synthesis at practical ratios: quality is maintained at 20:1 and, by PSNR/SSIM, remains close to baseline up to 100:1 (T1n $27.66$\,dB, T2w $25.95$\,dB), degrading markedly only at the extreme 200:1 and 400:1 settings (T1n $24.65$/$20.94$\,dB, T2w $22.63$/$17.18$\,dB).

Near-lossless JPEG-LS preserves synthesis quality across most \texttt{NEAR} levels. Significant degradation only happens at very severe compression levels (Table~\ref{tab:jpegls_combined}). This confirms that JPEG-LS retains the fidelity required for multi-modal generative training even though, as shown above, its raw-slice reconstruction error grows faster than JPEG2000's. Concurrent work on frozen foundation-VAE reconstructions~\cite{chen2026foundationvae} of 3D body CT reports the same phenomenon, where compact, noise-suppressed representations preserve or improve generative downstream tasks.

% The key result is that aggressive input compression does \emph{not} degrade synthesis: at 20:1, quality is maintained, and it degrades only at the extreme 200:1 setting (T1n $\rightarrow27.66$\,dB, T2w $\rightarrow25.95$\,dB).

\begin{table}[t]
    \centering
    \caption{JPEG2000: PSNR and SSIM across compression ratios and MRI modalities. Ratio 0 is the uncompressed baseline; the best per column is bolded.}
    \label{tab:jpeg2000_compression}
    \vspace{-0.5em}
    \small
    \begin{tabular}{lcccc|cccc}
    \toprule
    \multirow{2}{*}{Ratio} & \multicolumn{4}{c}{PSNR} & \multicolumn{4}{c}{SSIM} \\
    \cmidrule(lr){2-5}\cmidrule(lr){6-9}
     & T1c & T1n & T2f & T2w & T1c & T1n & T2f & T2w \\
    \midrule
    0   & $26.73$ & $28.39$ & $25.18$ & $27.65$ & $0.932$ & $0.983$ & $0.956$ & $\mathbf{0.967}$ \\
    2   & $\mathbf{28.26}$ & $28.26$ & $\mathbf{26.24}$ & $\mathbf{28.24}$ & $0.942$ & $0.964$ & $0.932$ & $0.951$ \\
    5   & $27.62$ & $\mathbf{29.40}$ & $25.27$ & $27.12$ & $0.938$ & $0.964$ & $0.924$ & $0.947$ \\
    10  & $27.95$ & $29.29$ & $25.54$ & $28.12$ & $0.941$ & $0.965$ & $0.925$ & $0.948$ \\
    20  & $26.81$ & $29.18$ & $26.16$ & $27.17$ & $0.935$ & $\mathbf{0.986}$ & $0.928$ & $0.948$ \\
    50  & $27.56$ & $28.71$ & $25.02$ & $27.05$ & $\mathbf{0.948}$ & $0.985$ & $\mathbf{0.958}$ & $0.961$ \\
    100 & $27.10$ & $27.66$ & $24.70$ & $25.95$ & $0.944$ & $0.982$ & $0.954$ & $0.952$ \\
    200 & $23.25$ & $24.65$ & $23.07$ & $22.63$ & $0.879$ & $0.968$ & $0.937$ & $0.902$ \\
    400 & $19.30$ & $20.94$ & $21.29$ & $17.18$ & $0.852$ & $0.829$ & $0.814$ & $0.750$ \\
    \bottomrule
    \end{tabular}
    \vspace{-1em}
\end{table}

\begin{table}[!ht]
    \centering
    \caption{JPEG-LS compression: PSNR and SSIM across compression ratios and MRI modalities. The synthesis quality remains highly robust across all contrasts, demonstrating that near-lossless JPEG-LS preserves the fidelity required for multi-modal generative modeling; the best per column is bolded.}
    \vspace{-0.5em}
    \small
    \label{tab:jpegls_combined}
    \begin{tabular}{lcccc|cccc}
        \toprule
        & \multicolumn{4}{c}{PSNR} & \multicolumn{4}{c}{SSIM} \\
        \cmidrule(lr){2-5} \cmidrule(lr){6-9}
        NEAR level & T1c & T1n & T2f & T2w & T1c & T1n & T2f & T2w \\
        \midrule
        $0$  & $26.73$ & $28.39$ & $25.18$ & $\mathbf{27.65}$ & $0.932$ & $0.983$ & $0.956$ & $\mathbf{0.967}$ \\
        $1$  & $27.21$ & $28.89$ & $\mathbf{25.49}$ & $27.57$ & $0.938$ & $\mathbf{0.986}$ & $\mathbf{0.961}$ & $0.963$ \\
        $2$  & $25.56$ & $28.35$ & $24.97$ & $25.79$ & $0.912$ & $0.983$ & $0.954$ & $0.947$ \\
        $5$  & $26.69$ & $28.84$ & $25.38$ & $27.64$ & $0.926$ & $\mathbf{0.986}$ & $0.960$ & $0.965$ \\
        $10$ & $27.29$ & $\mathbf{28.90}$ & $25.08$ & $26.93$ & $0.944$ & $\mathbf{0.986}$ & $0.955$ & $0.958$ \\
        $20$ & $24.28$ & $27.99$ & $25.36$ & $26.95$ & $0.859$ & $0.982$ & $0.960$ & $0.955$ \\
        $50$ & $\mathbf{27.52}$ & $28.31$ & $25.17$ & $27.50$ & $\mathbf{0.945}$ & $0.984$ & $0.959$ & $0.965$\\
        $200$ & $25.99$ & $26.44$ & $20.93$ & $ 24.28$ & $0.9267$ & $0.9764$ & $0.8471$ & $0.9273$\\
        $255$ & $25.54$ & $25.96$ & $23.24$ & $24.39$ & $0.9160$ & $0.9388$ & $0.8923$ & $0.9133$\\
        \bottomrule
    \end{tabular}
    \vspace{-2em}
\end{table}

\begin{figure}[htbp]
    \centering
    % --- Panel A: JPEG-LS ---
    \vspace{-1em}
    % \textbf{Near-Lossless Compression: JPEG-LS}\par\vspace{0.2em}
    \includegraphics[width=0.85\linewidth]{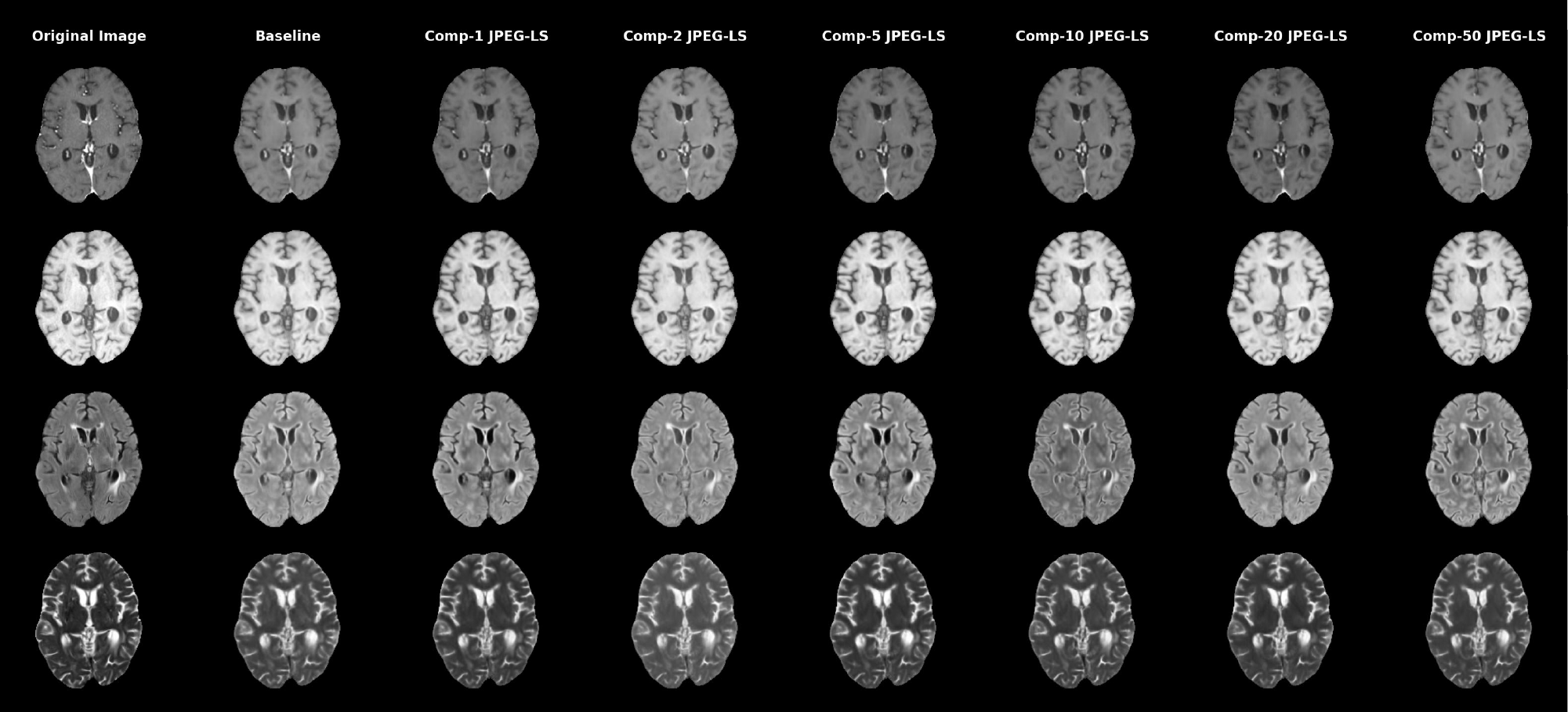} 
    
    % \vspace{0.1em}
    % --- Panel B: JPEG 2000 ---
    % \textbf{Lossy Compression: JPEG 2000}\par\vspace{0.2em}
    \includegraphics[width=0.85\linewidth]{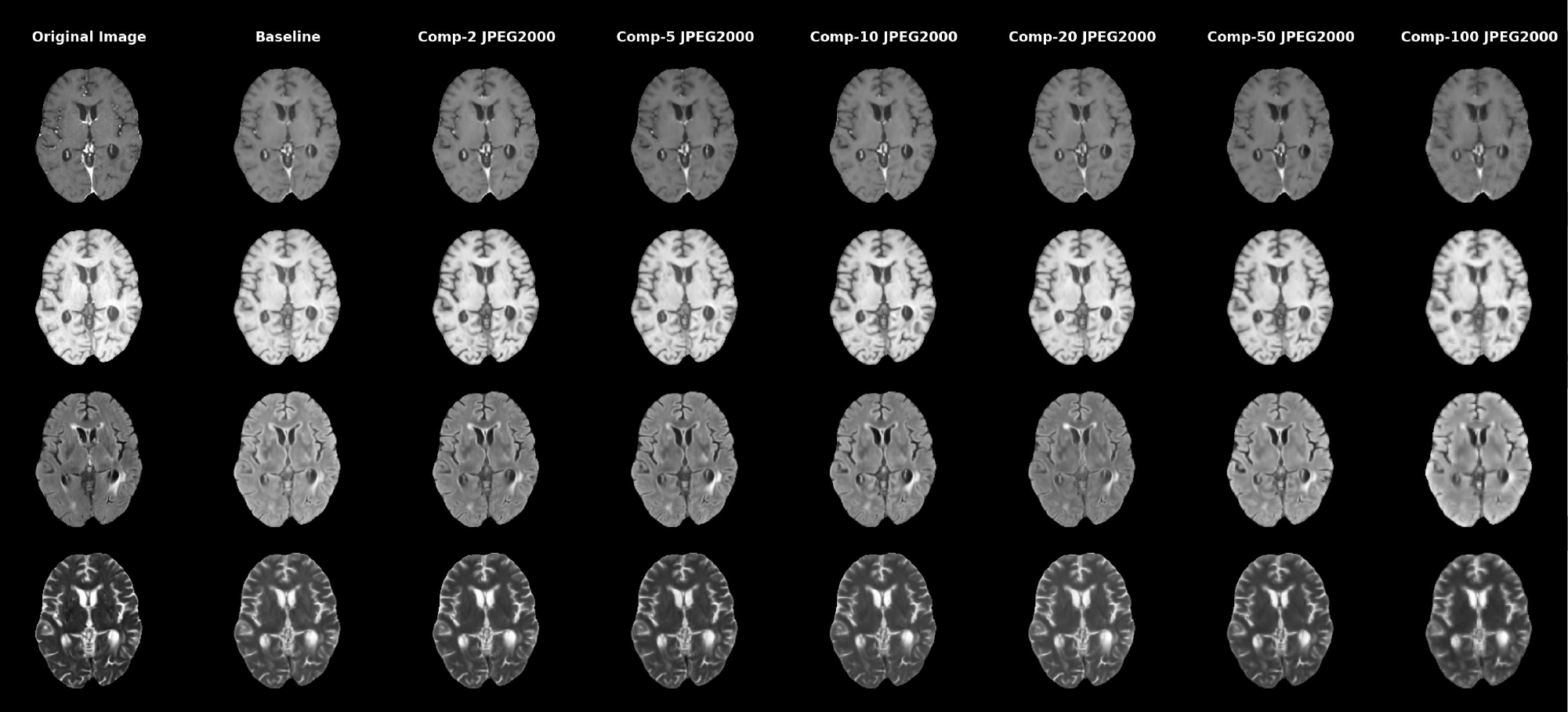} 
    \caption{WFM outputs are compared against the original images and a model trained on uncompressed data (Baseline). JPEG-LS (top): Near-lossless compression preserves fine anatomical structures, textures, and contrast boundaries across all four modalities. JPEG 2000 (bottom): While structural integrity is maintained at moderate ratios (up to 20:1), aggressive lossy compression settings (50:1, 100:1) introduce visible smoothing, blurring of high-frequency textures, and characteristic wavelet edge-ringing artifacts. Rows: top to bottom, T1c, T1n, T2f, and T2w modalities (Zoom for a good view).}
    \label{fig:qualitative_results}
    \vspace{-1em}
\end{figure}

\begin{figure}[!htbp]
    \centering
    \begin{minipage}[t]{0.49\textwidth}
        \centering
        \includegraphics[height=3.25cm]{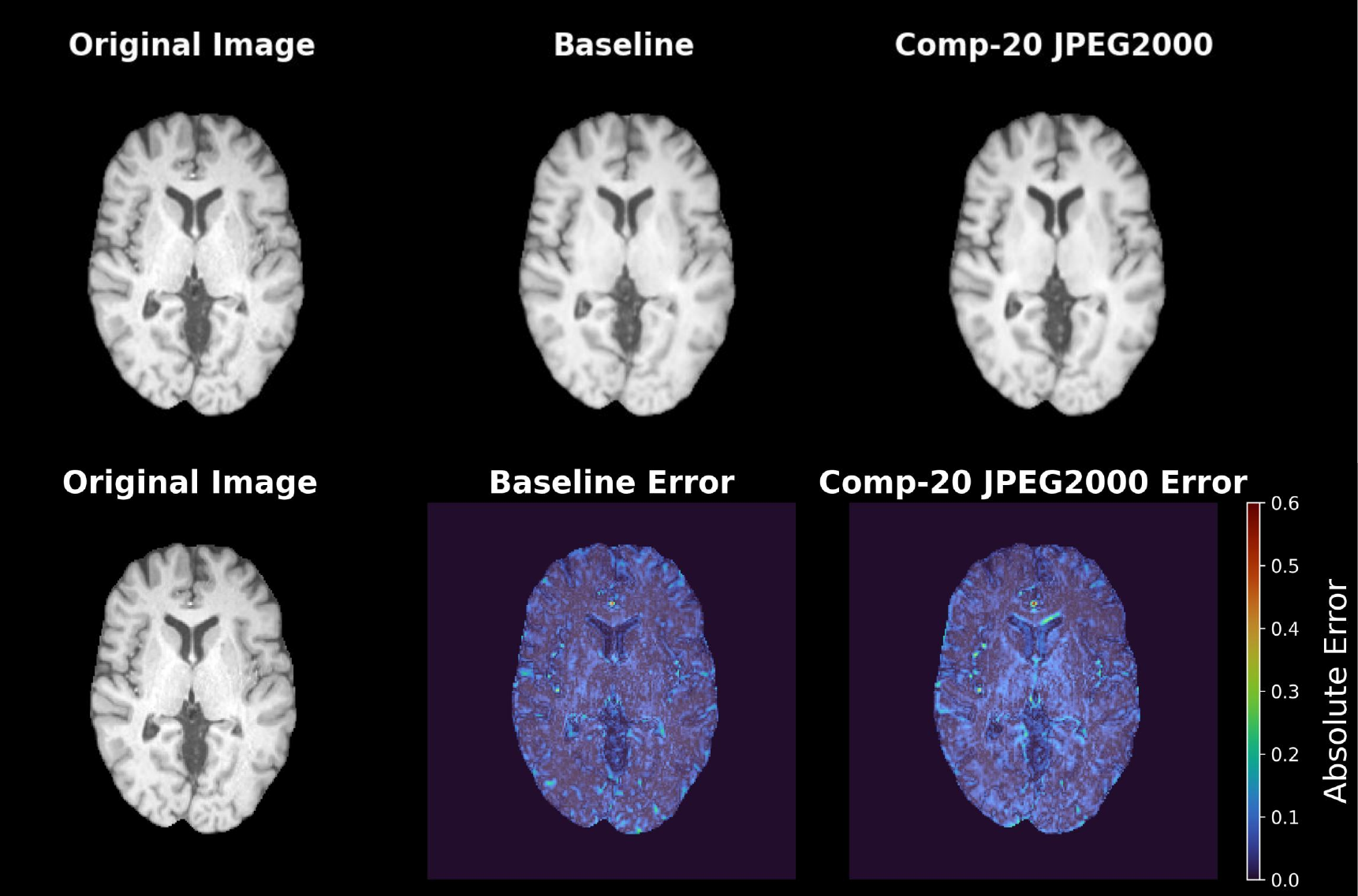}
        % \vspace{0.1em}
        % {\small (a) lowest PSNR JPEG2000}
    \end{minipage}
    \hfill
    \begin{minipage}[t]{0.49\textwidth}
        \centering
        \includegraphics[height=3.25cm]{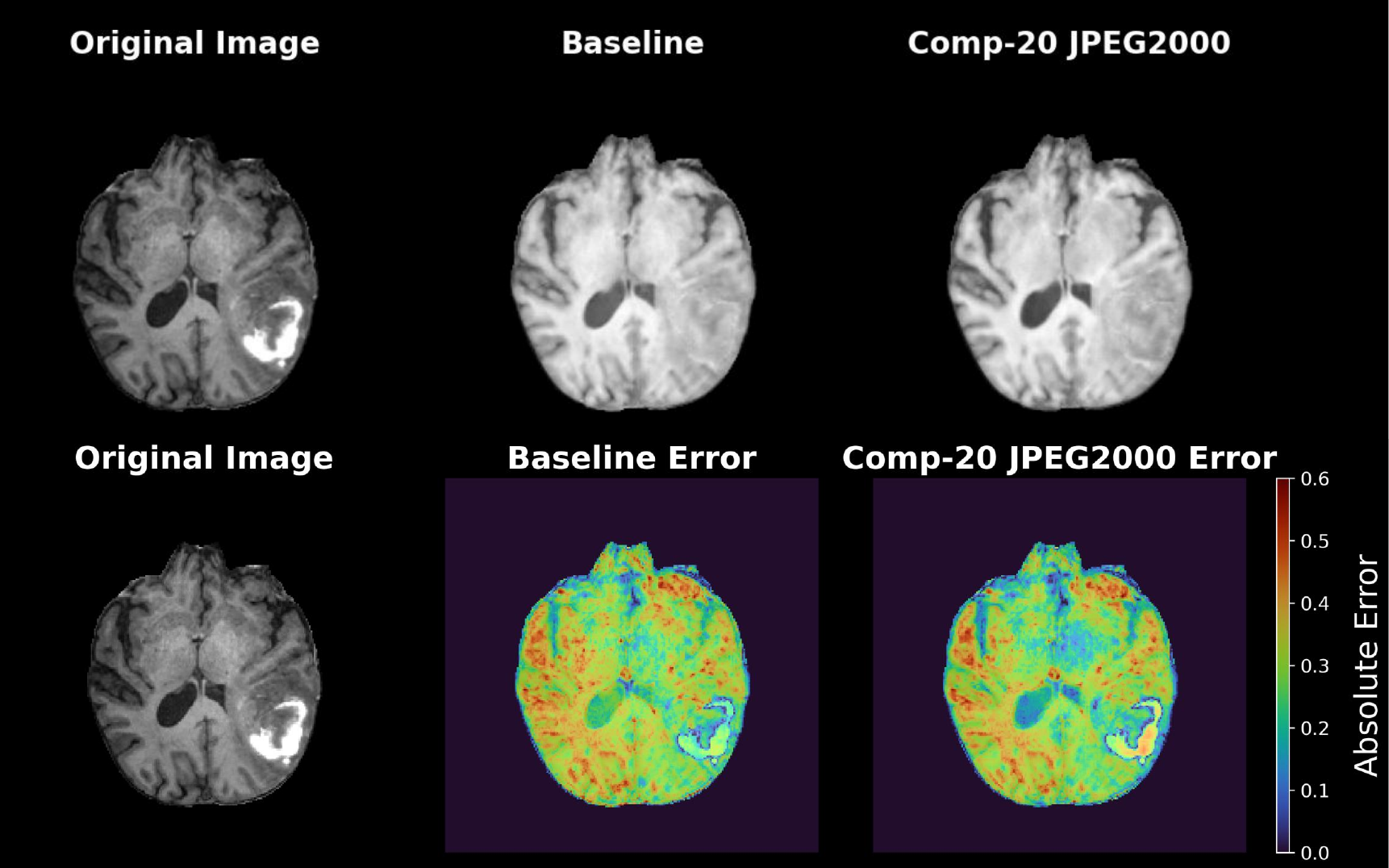}
        % {\small (b) highest PSNR JPEG2000}
    \end{minipage}
    \vspace{-1em}
    \caption{Qualitative comparison of MRI synthesis for best (left) and worst (right) case based on PSNR. From left to right: Uncompressed ground truth, baseline WFM synthesis (trained on lossless data), and Comp-20 synthesis (trained on 20:1 JPEG2000 compressed data). The Comp-20 model successfully preserves macroscopic tumor anatomy and contrast boundaries. However, subtle high-frequency texture smoothing is visible, consistent with the variance observed in fine-grained downstream segmentation tasks.}
    \label{fig:compression_results}
    \vspace{-1em}
\end{figure}

\paragraph{Qualitative evaluation:} Near-lossless JPEG-LS preserves fine anatomical structure and contrast boundaries across all four modalities, while JPEG2000 maintains structural integrity up to 20:1 and only exhibits visible smoothing and edge-ringing at 50:1--100:1 (Fig.~\ref{fig:qualitative_results}).
Fig.~\ref{fig:compression_results} contrasts the best- and worst-case validation subjects by PSNR: the 20:1 model preserves macroscopic tumor anatomy and contrast boundaries, with residual differences confined to subtle high-frequency texture, in the same regime that most affects fine-grained downstream tasks.

\paragraph{Inference speed:} Compression leaves inference throughput unchanged: both baseline and compressed models synthesize all four modalities in a single step at ${\sim}0.35$--$0.37$\,s per volume ($4\times188$ modality outputs in ${\sim}265$--$278$\,s total), since the codec acts only on stored data and not on the sampling path.

\begin{table}[!ht]
    \caption{Downstream task on the BraTS24 validation set using a state-of-the-art BraTS2025 algorithm, applied to volumes synthesized from compressed data. The metrics report agreement against pseudo-ground-truth masks from uncompressed NIfTI images; HD95 is averaged across cases with a defined value.}
    \label{tab:downstream_comp2}
    \centering
    % \vspace{-0.5em}
    \small
    \begin{tabular}{lcccccc}
        \toprule
        \multirow{2}{*}{Setting} & \multicolumn{2}{c}{Dice ($\uparrow$)} & \multicolumn{2}{c}{HD95 (mm) ($\downarrow$)} \\
        \cmidrule(lr){2-3}\cmidrule(lr){4-5}
         & WT & TC &  WT & TC \\
        \midrule
        JPEG2000 2:1   & $0.814$  & $0.238$ & $5.85$  & $22.84$ \\
        JPEG2000 50:1  & $0.777$ & $0.194$ & $7.72$ & $24.44$ \\
        JPEG-LS \texttt{NEAR}=1  & $0.802$ & $0.203$ &  $5.85$ & $23.32$ \\
        JPEG-LS \texttt{NEAR}=20 & $0.780$ & $0.185$ & $7.85$ & $26.97$ \\
        \bottomrule
    \end{tabular}
    \vspace{-2em}
\end{table}

\paragraph{Downstream task:} To probe whether synthesis from compressed images preserves clinically relevant structures, namely brain tumors, we performed whole-tumor segmentation using the winning solution from the BraTS2025 segmentation challenge~\cite{brats25}. We compared the resulting segmentations against masks derived from uncompressed inputs (Table~\ref{tab:downstream_comp2}). We observe that whole-tumor delineation is indeed well preserved (Dice $0.814$, 95th percentile Hausdorff distance $5.85$\, mm), whereas the smaller Tumor Core ($0.238$) region shows lower agreement and high variance. This gradient-robust macrostructure, fragile fine substructure, is consistent with the high-frequency smoothing observed qualitatively and marks the main limitation of compressed-data synthesis. These metrics quantify agreement between segmentations from compressed-data synthesis and uncompressed inputs; they isolate the compression effect but do not measure absolute accuracy against expert BraTS annotations. The low Tumor-Core agreement ($0.238$) indicates that small, high-frequency substructures are the primary casualty of aggressive compression, which we flag as the main clinical limitation.

\section{Discussion \& Conclusion}
We presented MRIComp4Flow, a pipeline for compressing 3D BraTS MRI using JPEG2000 and JPEG-LS, and for training a unified Wavelet Flow Matching synthesizer on the decompressed data. Across eight JPEG2000 ratios and six JPEG-LS levels, multi-modal synthesis quality remains comparable to training on lossless NIfTI up to 20:1 (a $12.9\times$ storage reduction) with unchanged sub-second inference; by PSNR/SSIM it holds up to 100:1 and collapses only at 200:1--400:1, while visible wavelet artifacts emerge earlier, at 50:1--100:1. Notably, at several intermediate ratios, we observe synthesis and downstream metrics that exceed the uncompressed baseline, suggesting that mild lossy compression can act as a beneficial denoising regularizer, an effect recently reported for frozen foundation-VAE reconstructions of 3D body CT~\cite{chen2026foundationvae}, and a promising avenue for deliberate exploitation. Building on this, a natural next step is to train generative models directly on the compressed latent representations rather than on decoded volumes, coupling storage efficiency with representation reuse.

\begin{credits}
\subsubsection{\ackname}
S.N. and A.MB. are supported by a TUM-IAS Fellowship. Funded by the Technical University of Munich -- Institute for Advanced Study (TUM-IAS), Germany.

\textbf{Dataset:} Data used in this publication were obtained as part of the Brain Tumor Segmentation (BraTS) Challenge project through Synapse ID:
syn53708249\footnote{\url{https://www.synapse.org/Synapse:syn53708249}}.
\end{credits}


\begin{thebibliography}{10}

\bibitem{bratscluster24}
Bakas, S., Baid, U., Rudie, J.D., Aboian, M., Anazodo, U., Calabrese, E., Conte, G.M., Fathi Kazerooni, A., Adewole, M., Alafif, M., Aljabar, P., Baig, S., Bergquist, T., Buchner, J.A., Correia de Verdier, M., Diaz-Pinto, A., Durrer, A., Ezhov, I., Foltyn-Dumitru, M., Gagnon, L., Hamamci, A., Hu, Q., Iglesias, J.E., Jiang, Z., Kofler, F., LaBella, D., Li, H.B., Linardos, A., Linguraru, M.G., Maleki, N., Moawad, A.W., Möller, H.E., Pati, S., Piraud, M., Rosier, M., Saluja, R., Schmick, A., Shirokikh, B., Steinbauer, F., Tahon, N.H., Wang, G., Wiestler, B., Zhang, J.:
The Brain Tumor Segmentation (BraTS) Cluster of Challenges (BraTS + Beyond-BraTS).
Zenodo.
\url{https://doi.org/10.5281/zenodo.10978907}
(2024)

\bibitem{balakrishnan2019voxelmorph}
Balakrishnan, G., Zhao, A., Sabuncu, M.R., Guttag, J., Dalca, A.V.:
VoxelMorph: A Learning Framework for Deformable Medical Image Registration.
IEEE Transactions on Medical Imaging \textbf{38}(8), 1788--1800 (2019).
\url{https://doi.org/10.1109/TMI.2019.2897538}

\bibitem{chen2026foundationvae}
Chen, Q., Ding, S., Gu, Y., Liu, N., Bian, J., Yuille, A., Zhou, Z., Fu, J.:
Foundation VAEs for 3D CT Reconstruction, Augmentation, and Generation.
arXiv preprint arXiv:2605.30893 (2026).

\bibitem{christopoulos2000jpeg2000}
Christopoulos, C., Skodras, A., Ebrahimi, T.:
The JPEG2000 Still Image Coding System: An Overview.
IEEE Transactions on Consumer Electronics \textbf{46}(4), 1103--1127 (2000).
\url{https://doi.org/10.1109/30.920468}

\bibitem{brats24}
Correia de Verdier, M., Saluja, R., Gagnon, L., LaBella, D., Baid, U., Tahon, N.H., Foltyn-Dumitru, M., Zhang, J., Alafif, M., Baig, S., Pati, S., Linardos, A., Adewole, M., Jiang, Z., Li, H.B., Conte, G.M., Calabrese, E., Maleki, N., Moawad, A.W., Anazodo, U., Fathi Kazerooni, A., Linguraru, M.G., Iglesias, J.E., Ezhov, I., Kofler, F., Piraud, M., Menze, B., Wiestler, B., Bakas, S.:
The 2024 Brain Tumor Segmentation (BraTS) Challenge: Glioma Segmentation on Post-treatment MRI.
arXiv:2405.18368 (2024).

\bibitem{danese2026flowlet}
Danese, D., Lombardi, A., Attimonelli, M., Fasano, G., Di Noia, T.:
FlowLet: Conditional 3D Brain MRI Synthesis using Wavelet Flow Matching.
arXiv preprint arXiv:2601.05212 (2026).

\bibitem{schelling2004jpegls}
Ebrahimi, F., Chamik, M., Winkler, S.:
JPEG vs. JPEG 2000: An Objective Comparison of Image Encoding Quality.
In: Rogowitz, B.E., Pappas, T.N. (eds.) Applications of Digital Image Processing XXVII.
Proceedings of SPIE, vol. 5558, pp. 300--308. SPIE (2004).
\url{https://doi.org/10.1117/12.560907}

\bibitem{friedrich2024wdm}
Friedrich, P., Wolleb, J., Bieder, F., Durrer, A., Cattin, P.C.:
WDM: 3D Wavelet Diffusion Models for High-Resolution Medical Image Synthesis.
In: Deep Generative Models. DGM4MICCAI 2024.
Lecture Notes in Computer Science, vol. 15320, pp. 11--21.
Springer, Cham (2024).
\url{https://doi.org/10.1007/978-3-031-72744-3_2}

\bibitem{isensee2021nnu}
Isensee, F., Jaeger, P.F., Kohl, S.A.A., Petersen, J., Maier-Hein, K.H.:
nnU-Net: A Self-configuring Method for Deep Learning-Based Biomedical Image Segmentation.
Nature Methods \textbf{18}(2), 203--211 (2021).
\url{https://doi.org/10.1038/s41592-020-01008-z}

\bibitem{karargyris2023medperf}
Karargyris, A., Umeton, R., Sheller, M.J., Aristizabal, A., George, J., Wuest, A., Pati, S., Baid, U., Chennubhotla, C., Ferrante, E., Edwards, B., Sharma, A., Bakas, S.: Federated Benchmarking of Medical Artificial Intelligence with MedPerf.
Nature Machine Intelligence \textbf{5}(8), 799--810 (2023). \url{https://doi.org/10.1038/s42256-023-00652-2}

\bibitem{Khader2023DM_MRi}
Khader, F., Müller-Franzes, G., Tayebi Arasteh, S., Han, T.,
Haarburger, C., Schulze-Hagen, M., Schad, P., Engelhardt, S.,
Baeßler, B., Foersch, S., Stegmaier, J., Kuhl, C.,
Nebelung, S., Kather, J.N., Truhn, D.:
Denoising Diffusion Probabilistic Models for 3D Medical Image Generation.
Scientific Reports \textbf{13}, 7303 (2023).
\url{https://doi.org/10.1038/s41598-023-34341-2}

\bibitem{brats25}
Kofler, F., Rosier, M., Astaraki, M., Baid, U., Möller, H.,
Buchner, J.A., Steinbauer, F., Oswald, E., de la Rosa, E.,
Ezhov, I., von See, C., Kirschke, J., Schmick, A., Pati, S.,
Linardos, A., Pitarch, C., Adap, S., Rudie, J., Correia de Verdier, M., Saluja, R., Calabrese, E., LaBella, D., Aboian, M., Moawad, A.W., Maleki, N., Anazodo, U., Adewole, M., Linguraru, M.G., Fathi Kazerooni, A., Jiang, Z., Conte, G.M., Li, H., Iglesias, J.E., Bakas, S., Wiestler, B., Piraud, M., Menze, B.: BraTS Orchestrator: Democratizing and Disseminating State-of-the-Art Brain Tumor Image Analysis. (2025).

\bibitem{kurmukov2024lossy}
Kurmukov, A., Zavolovich, B., Dalechina, A., Proskurov, V., Shirokikh, B.: The Effect of Lossy Compression on 3D Medical Images Segmentation with Deep Learning. (2024)

\bibitem{li2023brats}
Li, H.B., Conte, G.M., Hu, Q., Anwar, S.M., Kofler, F., Ezhov, I., Koen, V.L., Piraud, M., Diaz, M., Cole, B., Calabrese, E., Linardos, A., Rudie, J.D., Jiang, Z., Baid, U., Bakas, S.: The Brain Tumor Segmentation (BraTS) Challenge 2023: Brain MR Image Synthesis for Tumor Segmentation (BraSyn). (2023)

\bibitem{lipman2023flow}
Lipman, Y., Chen, R.T.Q., Ben-Hamu, H., Nickel, M., Le, M.: Flow Matching for Generative Modeling. In: International Conference on Learning Representations (ICLR). arXiv:2210.02747 (2023).

\bibitem{menze2015multimodal}
Menze, B.H., Jakab, A., Bauer, S., Kalpathy-Cramer, J., et al.:
The Multimodal Brain Tumor Image Segmentation Benchmark (BRATS).
IEEE Transactions on Medical Imaging \textbf{34}(10), 1993--2024 (2015). \url{https://doi.org/10.1109/TMI.2014.2377694}

\bibitem{ryoo2026performance}
Ryoo, J., Jung, Y., Khaliq, M.A., Zhang, W., Han, J., Lee, B.K.:
Performance Analysis and Optimization of 3D Generative Diffusion Models across GPU Architectures.
arXiv:2606.19365 (2026).

\bibitem{tur2026wfm}
Tur, Y., Stojkovic, M., Bagci, U.:
WFM: 3D Wavelet Flow Matching for Ultrafast Multi-Modal MRI Synthesis.
In: Proceedings of the 9th International Conference on Medical Imaging with Deep Learning (MIDL).
Proceedings of Machine Learning Research, vol. 315, pp. 3779--3796 (2026).

\bibitem{weinberger2000loco}
Weinberger, M.J., Seroussi, G., Sapiro, G.:
The LOCO-I Lossless Image Compression Algorithm: Principles and Standardization into JPEG-LS.
IEEE Transactions on Image Processing \textbf{9}(8), 1309--1324 (2000).

\bibitem{wyatt2022anoddpm}
Wyatt, J., Leach, A., Schmon, S.M., Willcocks, C.G.:
AnoDDPM: Anomaly Detection with Denoising Diffusion Probabilistic Models Using Simplex Noise.
In: Proceedings of the IEEE/CVF Conference on Computer Vision and Pattern Recognition Workshops (CVPRW), pp. 650--656 (2022).

\bibitem{predict_gbm}
Zimmer, L., Weidner, J., Balcerak, M., Kofler, F., Krupa, M., Ezhov, I., Cepeda, S., Zhang, R., Lowengrub, J., Menze, B., Wiestler, B.: PREDICT-GBM: A Multi-Center Platform to Advance Personalized Glioblastoma Radiotherapy Planning. arXiv:2509.13360v2 (2025).

\end{thebibliography}
\end{document}